\documentclass[10pt, a4paper]{article}
\usepackage{lrec}
\usepackage{multibib}
\newcites{languageresource}{Language Resources}
\usepackage{lipsum}
\usepackage{graphicx}
\usepackage{tabularx}
\usepackage{soul}
\usepackage{subfig}
\usepackage{epstopdf}
\usepackage[utf8]{inputenc}

\usepackage{hyperref}
\usepackage{xstring}

\usepackage{color}
\usepackage{xcolor}

\title{A Sentiment Analysis Dataset for Code-Mixed Malayalam-English}

\name{\begin{tabular}{c}Bharathi Raja Chakravarthi\(^1\), Navya Jose\(^2\), Shardul Suryawanshi\(^1\), \\ Elizabeth Sherly\(^2\), John P. McCrae\(^1\) \end{tabular}}
\address{ \(^1\)Insight SFI Research Centre for Data Analytics, Data Science Institute, National University of Ireland Galway \\ \{bharathi.raja,shardul.suryawanshi, john.mccrae\}@insight-centre.org\\
 \(^2\)Indian Institute of Information  Technology and Management-Kerala\\
\{navya.mi3, sherly\}@iiitmk.ac.in\\}

\abstract{
There is an increasing demand for sentiment analysis of text from social media which are mostly code-mixed. Systems trained on monolingual data fail for code-mixed data due to the complexity of mixing at different levels of the text.  However, very few resources are available for code-mixed data to create models specific for this data. Although much research in multilingual and cross-lingual sentiment analysis has used semi-supervised or unsupervised methods, supervised methods still performs better. Only a few datasets for popular languages such as English-Spanish, English-Hindi, and English-Chinese are available. There are no resources available for Malayalam-English code-mixed data. This paper presents a new gold standard corpus for sentiment analysis of code-mixed text in Malayalam-English annotated by voluntary annotators. This gold standard corpus obtained a Krippendorff's alpha above 0.8 for the dataset. We use this new corpus to provide the benchmark for sentiment analysis in Malayalam-English code-mixed texts.  
\\ \newline  }

\begin{document}
\maketitleabstract

\section{Introduction}
The Internet gave users the opportunities to express an opinion on any topic in the form of user reviews or comments. The comments are usually of an informal style, mostly in social media forums such as  Youtube, Facebook, and Twitter, which opens up the ground for mixing languages in the same conversation for multilingual communities. Some people with different linguistic backgrounds and cultures mark their impressions about a subject with the individual feeling in mixed language as not all are comfortable with a single language alone \cite{scotton1982possibility,tay1989code,suryawanshi-etal-2020-tamil-meme}. This unplanned switching between more than one language in the same conversation for the speaker's convenience is referred to as \textit{code-mixing} \cite{androutsopoulos2013code,chakravarthi2019comparison,chakravarthi-etal-2020-senti-tamil}.  Even though many languages have their own scripts, social media users use non-native script, usually Roman script, \cite{saint1987roman,rosowsky2010writing} for convenience in some part of the world, like India. This causes difficulties in finding the languages involved and also makes it hard to execute various existing natural language processing tasks, as these were developed for a single language \cite{bali-etal-2014-borrowing,ws-2014-approaches-code,solorio-etal-2014-overview}.

Malayalam is one of the Dravidian languages spoken in the southern region of India with nearly 38 million Malayalam speakers in India and other countries \cite{thottingal-2019-finite}. Malayalam is a deeply agglutinating language \cite{S18.125}. The Malayalam script is the Vatteluttu alphabet extended with symbols from the Grantha alphabet.   It is an alphasyllabary (abugida), a writing system that is partially “alphabetic” and partially syllable-based \cite{krishnamurti2003dravidian,lalitha-devi-2019-resolving,chakravarthi-etal-2019-multilingual}. Still, social media users use Roman script for typing due to it being easier to input. There is a lot of code-mixed data \cite{chakravarthi-etal-2019-multilingual} between Malayalam and English among the YouTube comments we surveyed. Monolingual datasets are available for Indian languages for various research aims \cite{agrawal-etal-2018-beating}. However, there are few attempts to make datasets for Malayalam code-mixed text. Thus traditional NLP tasks fail in this scenario due to the absence of a proper dataset. To create resources for a Malayalam-English code-mixed scenario, we collected comments of various Malayalam movie trailers from YouTube. Malayalam code-mixed sample text from the proposed dataset is shown below with the corresponding English glosses.
\begin{figure*}
  \includegraphics[width=\textwidth,height=4cm]{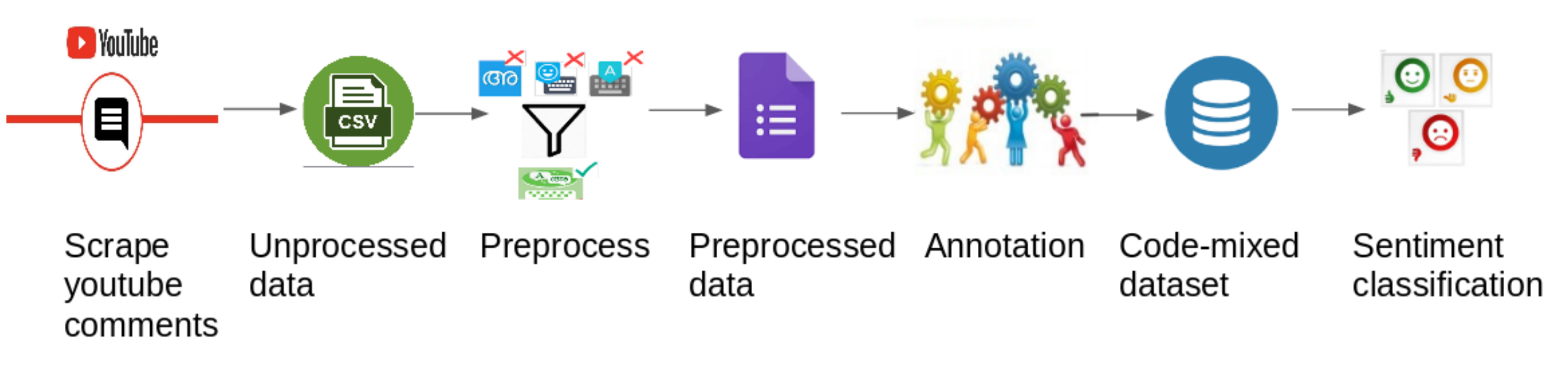}
  \caption{Data collection process.}
  \label{fig:flow}
\end{figure*}

\begin{itemize}

\item {\color{blue}Malayalam}-{\color{red}English}:  {\color{blue}\textit{Innaleyaaane kandath} {\color{red}super} {\color{blue}Padam.....ellarum} {\color{red}family}aaayi {\color{blue}poyi kananam} {\color{red}super} {\color{blue}abinayam}}

{\color{red}English: ``Watched yesterday only this super movie... everyone go and watch the movie with the family..super acting..''.}
\end{itemize}

The English words {\color{red}`super'} and {\color{red}`family'} intra-sententially code mixed \cite{barman-etal-2014-code} with the Malayalam language. Also, the word {\color{blue}`familyaayi'} is a new word combining both English and Malayalam, which is another kind of code-mixing called Intra-word switching that happens at the word level \cite{das-gamback-2014-identifying}. In this case, {\color{red}`with family'} is together said as a single word following Malayalam morphology. Although the main word {\color{red}`family’} is in English and as the sentence is in Malayalam, the new word takes Malayalam morphology. This comment can be considered as a positive comment from the viewer of the trailer of a Malayalam movie as it is clear that he enjoyed the movie and also recommend that movie to other viewers of the trailer.

\begin{itemize}
\item {\color{blue}Malayalam}-{\color{red}English}: \textit{{\color{blue}enthu oola {\color{red}trailer} aanu ithu.} {\color{red}poor dialogue delivery.}} 

{\color{red}English: ``What a useless trailer is this? Poor dialogue delivery.''}
\end{itemize}

This is an example of inter-sentential code-mixing \cite{barman-etal-2014-code}. {\color{blue}`Oola'} a slang word for {\color{red}`useless'} which is popular among the youth of Kerala.The viewer expressed strong dislike against the whole trailer and one aspect is {\color{red}`poor dialogue delivery'}. This comment has been marked as a negative comment as the disapproval of the trailer is evident.

Sentiment analysis is a topic of greater interest recently since business strategies can be enhanced with insights obtained from the opinion about the product or subject of interest from the users \cite{balage-filho-etal-2012-graphical,suryawanshi-etal-2020-meme}. As mentioned earlier, the greater part of comments in social media are code-mixed. The conducive nature of such platforms invits all users from different stratus of society to express their opinion about a subject with their own feeling. Hence it is true that the real sentiments about the subject can be extracted from the analysis of code-mixed data. Even with this massive enthusiasm for user-opinions, there is not much effort taken to analyse the sentiment of code-mixed content in under-resourced languages. The contribution of this paper is that we release the gold-standard code-mixed dataset for Malayalam-English annotated for sentiment analysis and provide comprehensive results on popular classification methods. To the best of our knowledge, this is the first code-mixed dataset for Malayalam sentiment analysis. Our code implementing these models along with the dataset is available freely for research purposes\footnote{https://github.com/bharathichezhiyan/MalayalamMixSentiment}.

\section{Related Work}
The sentiment analysis task has become increasingly important due to the explosion of social media, and extensive research has been done for sentiment analysis of monolingual corpora such as English \cite{10.1145/1014052.1014073,Wiebe2005,jiang-etal-2019-challenge}, Russian \cite{rogers-etal-2018-rusentiment}, German \cite{cieliebak-etal-2017-twitter}, Norwegian \cite{maehlum-etal-2019-annotating} and Indian languages \cite{agrawal-etal-2018-beating,priya-etal-2020-senti-comparative}.  

There have been two traditional approaches to solve sentiment analysis problem such as lexicon-based, and machine learning approaches \cite{Habimana2019}. With the increasing popularity of lexicons in the field of sentiment analysis since 1966, new lexicons namely WordNet \cite{wordnet}, WordNet-Affect \cite{Valitutti04wordnet-affect:an,chakravarthi2018improving,chakravarthi-etal-2019-wordnet}, SentiNet \cite{sentinet}, and SentiWordNet \cite{Esuli2006sentiwordnet} were primarily used. Although being famous for their simplicity, both traditional machine learning and lexicon-based approaches are not efficient when applied on user-generated data, due to the dynamic nature of such data. This is where deep learning approaches take the spotlight for being efficient in adapting to dynamic user-generated data. In the advent of transfer learning, GloVe \cite{pennington2014glove}, Word2Vec \cite{word2vec}, fastText \cite{bojanowski-etal-2017-enriching} comes with their pros and cons. 

Malayalam \cite{nair2014sentima,sarkar2015sentiment,se2015amrita,se2016predicting,mouthami2013sentiment} has official status in India and other countries. Several research activities on sentiment analysis and events are focused on Malayalam due to their population and use of this language. However, sentiment analysis on Malayalam-English is very low, and data are not easily available for the research. Code-mixed data contains informal language with numerous accidental, deliberate errors, mixing of language and grammatical mixing, which makes previous corpora and methods less suitable to train a model for sentiment analysis in code-mixed data. 

In the past few years, there have been increasing efforts on a variety of task using code-mixed text. However, the number of a freely available code-mixed dataset \cite{chakravarthi2016,chakravarthi-code-mix-survey} are still limited in number, size, availability. For few languages, such as English-Hindi \cite{joshi-etal-2016-towards,patra2018sentiment,chakravarthi-code-mix-ruba-ne}, English-Spanish \cite{solorio-etal-2014-overview} , Chinese-English \cite{lee-wang-2015-emotion}, and English-Bengali \cite{patra2018sentiment} datasets are available for research.  There are no dataset for Malayalam-English, so inspired by \newcite{severyn-etal-2014-opinion} we collected and created a code-mixed dataset from YouTube. 
We provided a use case of the code-mixed Malayalam-English dataset by laying down the baselines which make the use of state of the art techniques such as Dynamic Meta-Embeddings {DME} \cite{kiela-etal-2018-dynamic}, Contextualized DME {CDME} \cite{kiela-etal-2018-dynamic}, 1D Dimensional Convolution {1DConv} \cite{zhou2016text}, and Bidirectional Encoder Representations for Transformers {BERT} \cite{devlin2018bert}.

\section{Corpus Creation and Annotation}
Our goal was to create a code-mixed dataset for Malayalam-English and to ensure that enough data are available for research purposes.  We used \textit{youtube-comment-scraper tool} \footnote{https://github.com/philbot9/youtube-comment-scraper} to download the comments from YouTube. First, we collected 116,711 sentences for Malayalam from YouTube post comments. We collect the comments from the movie trailers of 2019 based on the YouTube search results for keyword "Malayalam movie 2019". Many of the comments that we downloaded were either fully in English or mixed. Therefore, we filtered out non-code-mixed corpus bases on language identification at comment level with the \textit{langdect library} \footnote{https://pypi.org/project/langdetect/}. That is if the comment is fully in one language than we discarded that comment since monolingual resources are available for these languages. Comments in Malayalam script was also discarded. We preprocessed the comments by removing the emoji's, and sentence length longer than 15 or less than 5 words since sentence more than 15 words will be difficult for annotators. After cleaning, we got 6,738 sentences for Malayalam-English code-mixed post comments.
\subsection{Annotation Setup}
For annotation, we adopted the approach taken by \newcite{mohammad-2016-practical} and each sentence was annotated by a minimum of three annotators according to the following schema: 
\begin{itemize}
    \item \textbf{Positive state:} There is an explicit or implicit clue in the text suggesting that the speaker is in a positive state, i.e., happy, admiring, relaxed, and forgiving.
    \item \textbf{Negative state:} There is an explicit or implicit clue in the text suggesting that the speaker is in a negative state, i.e., sad, angry, anxious, and violent. 
    \item \textbf{Mixed feelings:}  There is an explicit or implicit clue in the text suggesting that the speaker is experiencing both positive and negative feeling: Comparing two movies 
    \item \textbf{Neutral state:} There is no explicit or implicit indicator of the speaker’s emotional state: Examples are asking for like or subscription or questions about the release date or movie dialogue. This state can be considered as a neutral state.
    \item \textbf{Not in intended language:} For Malayalam if the sentence does not contain Malayalam then it is not Malayalam.
\end{itemize}
We anonymized sensitive elements that may result in the problem of confidentiality in the YouTube comments.   We created Google Forms, in which we collected the annotator's email so the annotator can annotate only once. We collected gender, education and medium of schooling information to know the diversity of the annotators, and we informed the annotators about the use of the data for finding the diversity of annotators. The annotators were given a choice to quit the annotation whenever they are uncomfortable with annotation.  Each Google Form has to set contain a maximum of 100 sentences. The annotation of each corpus was performed in three phases. First, each sentence was annotated by two annotators. The second step, the data were collected if both annotators agreed, in the case of conflict, a third annotator annotated the sentence. In the third step, if all the three annotators did not agree, then two more annotators annotated the sentences.

\subsection{Annotators}
Once the Google form was ready, we sent it out to an equal number of male and females to annotate. In the end, six annotators volunteered to annotate all of who are  Malayalam-English bilingual proficiency and ready to take up the task seriously. From Table \ref{tab:annotators}, we can see that four female and two male voluntarily annotated our forms. All of them were postgraduates. Though among the annotators only one did schooling in native (Malayalam) medium and others in English medium, we ensured it would not affect the task as all of them are fully proficient at using this language. 
\begin{table}[t] 
\begin{center} 
\begin{tabular}{|l|l|l|l|}
\hline
& & Malayalam \\
\hline
Gender & Male & 2  \\
& Female & 4 \\
\hline
Highest Education & Undegraduate & 0 \\
& Graduate & 0  \\
& Postgraduate & 6 \\
\hline
Medium of Schooling & English & 5  \\
& Native & 1 \\
\hline
Total && 6 \\
\hline
\end{tabular}
\caption{Annotators } 
\label{tab:annotators} 
\end{center} 
\end{table}

\subsection{Corpus Statistics}
Table \ref{tab:corp_stat} shows the corpus statistics of Malayalam-English code-mixed dataset. As is shown, this huge corpus at the end has 70,075 tokens, where 19,992 are unique. There are 6,739 comments and 7,743 distinct sentences in our code-mixed sentiment dataset. On average, there are ten tokens per sentence, and there is at least one sentence per post.

As mentioned before, the whole data has been categorized into five groups viz: positive, negative, neutral, mixed feeling, non-Malayalam. The distribution of data each category is detailed in Table \ref{tab:data_distribution}. Out of 6,739 posts, 2,811 comments have a positive polarity which is the most frequent category here. If there is no indication of the speaker’s emotional state about the subject in the post, the post belongs to a neutral state which is the second-largest category with 1,903 posts here. This may be due to the increasing trend of asking for likes to their comments by the users. We split the corpus retaining  20 percentage that is 1,348 for test, 10 percentage for validation that is 674 for validation, and remaining for training.

\begin{table*}[t] 
\begin{center} 
\begin{tabular}{|l|r|}
\hline
Language pair & Malayalam-English  \\
\hline
Number of Tokens & 70,075 \\
Vocabulary Size & 19,992 \\
Number of Posts & 6,739\\
Number of Sentences & 7,743 \\
Average Sentence Length & 10 \\
Average number of sentences per post & 1 \\

\hline
\end{tabular} 
\caption{Corpus statistic of Malayalam-English Data } 
\label{tab:corp_stat} 
\end{center} 
\end{table*}

\begin{table}[t] 
\begin{center} 
\begin{tabular}{|l|r|}
\hline
Class & Malayalam-English\\
\hline
Positive & 2,811  \\
Negative & 738 \\
Mixed feelings & 403  \\
Neutral state & 1,903 \\
Non-Malayalam & 884  \\
\hline
Total & 6,739 \\
\hline
\end{tabular} 

\caption{Data Distribution} 
\label{tab:data_distribution} 
\end{center} 
\end{table}

\subsection{Inter Annotator Agreement}

While labelling, the corpus linguist has to decide independently to which category the comment to be added following the guidelines provided strictly. It could be inferred that the guidelines for annotation were clearly understood by all the annotators if they made the same annotations freely. Because of this existence of more than one annotator to label the same set of data, it is necessary to have a metric to compare those annotation qualities. This motivates the use of inter-annotator agreement which says how good the annotation decisions made by those multiple annotators on the same dataset are. A high score on this statistical metric does not mean the annotations are accurate, but it shows the homogeneity of agreement among the corpus linguists about the category. In other words, high inter-annotator agreement implies guidelines are clear, and interpretations are correct.

Though computationally complex, we used \textbf{Krippendorff's alpha $(\alpha)$} a prominent method among the numerous approaches developed to measure the degree of agreement between annotators. Krippendorff's alpha $(\alpha)$ is more relevant in our case as it is not affected by missing data, takes care of varying sample sizes, categories, numbers of raters and can also be employed to any measurement levels like nominal, ordinal, interval, ratio. Since more than two people have done the annotation task here and the same peoples annotate not all sentences, Krippendorff's alpha $(\alpha)$ fits here more. We used \textit{nltk}\footnote{https://www.nltk.org/} for calculating Krippendorff's alpha $(\alpha)$. Our annotation produced an agreement of 0.890 using nominal metric and 0.911 using interval metric.

   


\section{Difficult Examples}
While annotating, a few of the comments were ambiguous about sensing the right feelings from the viewers. Hence the task of annotation for sentiment analysis seemed difficult. The problems include \textbf{the comparison of the movie with movies of same or other industries, expression of opinion of different aspects of the movie in the same sentence}. Below shows a few examples of such comments and detailed how we resolved those issues.
\begin{itemize}

\item {\color{blue}\textit{Kanditt Amala Paul Aadai Tamil mattoru {\color{red}version} aanu ennu thonnunu}}

``It looks like another version of amala paul's Tamil movie aadai''.

Here the viewer doubts the Malayalam movie 'Helen' is similar to the Tamil movie `Aadai'.Though that movie `Aadai' was a positively reviewed movie by viewers and critics, we cannot generalize and assume this comment also as positive only because of this comparison. Hence we add it to the category of `mixed feeling'.

\item {\color{blue}\textit{Evideo oru {\color{red}Hollywood story} varunnilleee. Oru DBT.}}

``Somewhere there is a Hollywood storyline...one doubt."

This is also a comparison comment of that same movie `Helen' mentioned above. Nevertheless, here the difference is that it is compared with the whole Hollywood standard, which is accepted worldwide. Hence it is marked as a positive comment.
\item {\color{blue}\textit{{\color{red}Trailer} pole nalla {\color{red}story} undayal mathiyarinu.}}

``It was good enough to have a good story like the trailer".

Here viewer mentioned about two aspects of that movie viz: `trailer' and `story'. He appreciates the trailer but at the same time doubt about the story. This comment we considered as a positive comment as it is clear that he enjoyed the trailer and also shows strong optimism for that particular movie.
\end{itemize}

\begin{table*}[!ht]
\centering
\begin{tabular} {c|cccc|cccc}
\hline
 	& \multicolumn{4}{c|}{\textbf{LR}} & \multicolumn{4}{c}{\textbf{SVM}}\\
	\hline
    &Precision & Recall & f1-score &support &Precision & Recall & f1-score &support\\
    \hline
	Mixed feelings & 0.59 & 0.23 & 0.33 & 70 & 0.00 & 0.00 & 0.00 & 70\\
	Negative & 0.70 & 0.45 & 0.55 & 138 & 0.00 & 0.00 & 0.00 & 138\\
	Positive & 0.68 & 0.83 & 0.75 & 565 & 0.00 & 0.00 & 0.00 & 565\\
	Non-Malayalam & 0.69 & 0.58 & 0.63 & 177 & 0.13 & 1.00 & 0.23 & 177\\
	Neutral & 0.65 & 0.65 & 0.65 & 398 & 0.00 & 0.00 & 0.00 & 398\\
	\hline
	macro avg & \textbf{0.66} & 0.55 & 0.58 & 1348 & 0.03 & 0.20 & 0.05 & 1348\\
	weighted avg & 0.67 & 0.67 & 0.66 & 1348 & 0.02 & 0.13 & 0.03 & 1348\\
 \hline
 	& \multicolumn{4}{c|}{\textbf{DT}} & \multicolumn{4}{c}{\textbf{RF}}\\
	\hline
	Mixed feelings & 0.21 & 0.19 & 0.20 & 70 & 0.50 & 0.03 & 0.05 & 70\\
	Negative & 0.60 & 0.44 & 0.51 & 138 & 0.75 & 0.37 & 0.50 & 138\\
	Positive & 0.62 & 0.67 & 0.65 & 565 & 0.62 & 0.83 & 0.71 & 565\\
	Non-Malayalam & 0.42 & 0.57 & 0.49 & 177 & 0.61 & 0.60 & 0.61 & 177\\
	Neutral & 0.56 & 0.46 & 0.51 & 398 & 0.64 & 0.56 & 0.60 & 398\\
	\hline
	macro avg & 0.48 & 0.47 & 0.47 & 1348 & 0.62 & 0.48 & 0.49 & 1348\\
	weighted avg & 0.55 & 0.55 & 0.55 & 1348 & 0.63 & 0.63 & 0.61 & 1348\\
	\hline
 	& \multicolumn{4}{c|}{\textbf{MNB}} & \multicolumn{4}{c}{\textbf{KNN}}\\
	\hline
	Mixed feelings & 0.00 & 0.00 & 0.00 & 70 & 0.50 & 0.01 & 0.03 & 70\\
	Negative & 0.93 & 0.10 & 0.18 & 138 & 0.75 & 0.02 & 0.04 & 138\\
	Positive & 0.54 & 0.93 & 0.68 & 565 & 0.73 & 0.11 & 0.19 & 565\\
	Non-Malayalam & 0.84 & 0.21 & 0.34 & 177 & 0.16 & 0.63 & 0.25 & 177\\
	Neutral & 0.66 & 0.52 & 0.58 & 398 & 0.34 & 0.47 & 0.39 & 398\\
	\hline
	macro avg & 0.60 & 0.35 & 0.36 & 1348 & 0.50 & 0.25 & 0.18 & 1348\\
	weighted avg & 0.63 & 0.58 & 0.52 & 1348 & 0.53 & 0.27 & 0.23 & 1348\\
 \hline
 	& \multicolumn{4}{c|}{\textbf{DME}} & \multicolumn{4}{c}{\textbf{CDME}}\\
	\hline
	Mixed feelings & 0.11 & 0.06 & 0.07 & 70 & 0.08 & 0.11 & 0.10 & 70 \\
	Negative & 0.20 & 0.12 & 0.15 & 138 & 0.24 & 0.12 & 0.16 & 138 \\
	Positive & 0.59 & 0.44 & 0.50 & 565 & 0.57 & 0.52 & 0.54 & 565 \\
	Non-Malayalam & 0.23 & 0.62 & 0.34 & 177 & 0.29 & 0.76 & 0.42 & 177 \\
	Neutral & 0.52 & 0.44 & 0.47 & 398 & 0.67 & 0.35 & 0.46 & 398 \\
	\hline
	macro avg & 0.33 & 0.33 & 0.31 & 1348 & 0.37 & 0.37 & 0.34 & 1348 \\
	weighted avg & 0.45 & 0.41 & 0.41 & 1348 & 0.51 & 0.44 & 0.44 & 1348 \\
	 \hline
 	& \multicolumn{4}{c|}{\textbf{1DConv}} & \multicolumn{4}{c}{\textbf{BERT}}\\
	\hline
	Mixed feelings & 0.26 & 0.26 & 0.26 & 70 & 0.00 & 0.00 & 0.00 & 70 \\
	Negative & 0.56 & 0.33 & 0.41 & 138 & 0.57 & 0.55 & 0.56 & 138 \\
	Positive & 0.70 & 0.74 & 0.72 & 565 & 0.83 & 0.87 & 0.85 & 565 \\
	Non-Malayalam & 0.62 & 0.75 & 0.68 & 177 & 0.87 & 0.93 & 0.90 & 177 \\
	Neutral & 0.61 & 0.60 & 0.61 & 398 & 0.73 & 0.79 & 0.76 & 398 \\
	\hline
	macro avg & 0.55 & 0.54 & 0.54 & 1348 & 0.60 & \textbf{0.63} & \textbf{0.61} & 1348 \\
	weighted avg & 0.63 & 0.63 & 0.63 & 1348 & \textbf{0.73} & \textbf{0.78} & \textbf{0.75} & 1348 \\  
\hline
\end{tabular}
\caption{Precision, recall, F1-score and support for Logistic regression (LR), Support vector machine (SVM), Decision tree (DT), Random Forest (RF), Multinomial Naive Bayes (MNB), K-nearest neighbours (KNN), DME, CDME, 1DConv, BERT.}
\label{tab:whole}
\end{table*}

\section{Benchark Systems}
Traditional machine learning algorithm such as Logistic regression (LR), Support vector machine (SVM), Decision tree (DT), Random Forest (RF), Multinomial Naive Bayes (MNB), K-nearest neighbours (KNN) have been used on the newly annotated English-Malayalam dataset to show the insights about the dataset. The input features are the Term Frequency Inverse Document Frequency (TF-IDF). This approach makes these models trained only on this dataset without taking any pre-trained embeddings.

We also show the result in the deep learning-based models. This is due to the dynamic nature of the data as it is hard to derive a pattern just by using the handcrafted features, which later could be feed inside the algorithms such as logistic regression (LR), support vector machines (SVM). To provide a simple baseline, we implemented four models, which includes Dynamic Meta-Embeddings {DME} \cite{kiela-etal-2018-dynamic}, Contextualized DME {CDME} \cite{kiela-etal-2018-dynamic}, 1D Dimensional Convolution {1DConv} \cite{zhou2016text}, Bidirectional Encoder Representations for Transformers {BERT} \cite{devlin2018bert}. 

We evaluated our dataset based on the precision, recall and F-score of these baselines. We used \textit{sklearn} \footnote{https://scikit-learn.org/}, the micro average is calculated globally by counting the total true positives, false negative and false positive. The macro average
compute the metric independently for each class and then take the unweighted mean. The macro average does not take imbalance into account. A weighted average calculated for each label like macro, and find their average weighted by support. For our test, there are 2,075 positive examples, 424 negative, 173 neutral, 377 mixed feelings, and 100 non-Malayalam examples. This is a variation of macro to include label imbalance. This may cause F-score not be between precision and recall.

We combined fastText \cite{bojanowski2017enriching} and word2vec \cite{mikolov2013efficient} in DME and CDME baselines. The fastText and word2vec were trained on our code-switched dataset. 
In DME, we are combining the mentioned embeddings by doing a weighted sum. On the otherhand CDME is using self-attention based mechanism on top of DME to make the embeddings context-dependent. 1DConv makes use of a 1D convolution filter to represent each word with the context of the neighbouring word in the range of the kernel. In this convolutional neural network (CNN) \cite{kalchbrenner2014convolutional} approach, we are trying to capture the standout features from the text.  BERT makes use of encoder-decoder architecture with an attention mechanism which increases the flexibility to read sequence both (left to right and vice versa) ways. 

From the results shown in the Table \ref{tab:whole}, all the machine learning algorithms succeed in classifying all the classes except SVM. A recall of 1.00 and precision around 0.13 for non-Malayalam class shows that all the classes have been labelled as non-Malayalam irrespectively. Other than SVM, LR, DT, RF shows  considerable macro average score for precision, recall and F1-score. However, MNB and KNN achieve higher macro-averaged precision at the expense of the lower recall values. 
As mentioned earlier, deep learning models are using pre-trained embeddings. The use of fastText in combination with word2vec for DME and CDME gives both local as well as global context. 1DConv shows the better macro-averaged score in precision, recall and F1-score, BERT, on the other hand, fails to identify ``Mixed feeling'' class. However, DME and CDME succeed in identifying all the classes.

\section{Conclusion}
In this paper, we have presented the Malayalam-English corpus a code-mixed corpus of YouTube comments annotated for sentiment analysis. This annotation project aims to allow researches to enable research on code-mixed sentiment analysis, as well as provide useful data for code-mixed research. We also provide an inter-annotator agreement score in terms of Kripendorff's alpha and baseline results, as well as making the corpus available to the research community.
\section{Acknowledgments}
This publication has emanated from research supported in part by a research grant from Science Foundation Ireland (SFI) under Grant Number SFI/12/RC/2289 (Insight), SFI/12/RC/2289$\_$P2 (Insight$\_$2), co-funded by the European Regional Development Fund as well as by the EU H2020 programme under grant  agreements 731015 (ELEXIS-European Lexical Infrastructure), 825182 (Prêt-à-LLOD), and Irish Research Council grant IRCLA/2017/129 (CARDAMOM-Comparative Deep Models of Language for Minority and Historical Languages).
\section{Bibliographical References}
\bibliographystyle{lrec}
\bibliography{lrec2020W-xample-kc}

\begin{thebibliography}{}

\bibitem[\protect\citename{Agrawal \bgroup et al.\egroup
  }2018]{agrawal-etal-2018-beating}
Agrawal, R., Chenthil~Kumar, V., Muralidharan, V., and Sharma, D.
\newblock (2018).
\newblock No more beating about the bush : A step towards idiom handling for
  {I}ndian language {NLP}.
\newblock In {\em Proceedings of the Eleventh International Conference on
  Language Resources and Evaluation ({LREC} 2018)}, Miyazaki, Japan, May.
  European Language Resources Association (ELRA).

\bibitem[\protect\citename{Androutsopoulos}2013]{androutsopoulos2013code}
Androutsopoulos, J.
\newblock (2013).
\newblock Code-switching in computer-mediated communication.
\newblock {\em Pragmatics of computer-mediated communication}, pages 667--694.

\bibitem[\protect\citename{Balage~Filho \bgroup et al.\egroup
  }2012]{balage-filho-etal-2012-graphical}
Balage~Filho, P.~P., Brun, C., and Rondeau, G.
\newblock (2012).
\newblock A graphical user interface for feature-based opinion mining.
\newblock In {\em Proceedings of the Demonstration Session at the Conference of
  the North {A}merican Chapter of the Association for Computational
  Linguistics: Human Language Technologies}, pages 5--8, Montr{\'e}al, Canada,
  June. Association for Computational Linguistics.

\bibitem[\protect\citename{Bali \bgroup et al.\egroup
  }2014]{bali-etal-2014-borrowing}
Bali, K., Sharma, J., Choudhury, M., and Vyas, Y.
\newblock (2014).
\newblock {``}{I} am borrowing ya mixing ?'' an analysis of {E}nglish-{H}indi
  code mixing in {F}acebook.
\newblock In {\em Proceedings of the First Workshop on Computational Approaches
  to Code Switching}, pages 116--126, Doha, Qatar, October. Association for
  Computational Linguistics.

\bibitem[\protect\citename{Barman \bgroup et al.\egroup
  }2014]{barman-etal-2014-code}
Barman, U., Das, A., Wagner, J., and Foster, J.
\newblock (2014).
\newblock Code mixing: A challenge for language identification in the language
  of social media.
\newblock In {\em Proceedings of the First Workshop on Computational Approaches
  to Code Switching}, pages 13--23, Doha, Qatar, October. Association for
  Computational Linguistics.

\bibitem[\protect\citename{Bojanowski \bgroup et al.\egroup
  }2017a]{bojanowski-etal-2017-enriching}
Bojanowski, P., Grave, E., Joulin, A., and Mikolov, T.
\newblock (2017a).
\newblock Enriching word vectors with subword information.
\newblock {\em Transactions of the Association for Computational Linguistics},
  5:135--146.

\bibitem[\protect\citename{Bojanowski \bgroup et al.\egroup
  }2017b]{bojanowski2017enriching}
Bojanowski, P., Grave, E., Joulin, A., and Mikolov, T.
\newblock (2017b).
\newblock Enriching word vectors with subword information.
\newblock {\em Transactions of the Association for Computational Linguistics},
  5:135--146.

\bibitem[\protect\citename{Chakravarthi \bgroup et al.\egroup
  }2018]{chakravarthi2018improving}
Chakravarthi, B.~R., Arcan, M., and McCrae, J.~P.
\newblock (2018).
\newblock {Improving Wordnets for Under-Resourced Languages Using Machine
  Translation}.
\newblock In {\em Proceedings of the 9th Global WordNet Conference}. The Global
  WordNet Conference 2018 Committee.

\bibitem[\protect\citename{Chakravarthi \bgroup et al.\egroup
  }2019a]{chakravarthi2019comparison}
Chakravarthi, B.~R., Arcan, M., and McCrae, J.~P.
\newblock (2019a).
\newblock Comparison of different orthographies for machine translation of
  under-resourced dravidian languages.
\newblock In {\em 2nd Conference on Language, Data and Knowledge (LDK 2019)}.
  Schloss Dagstuhl-Leibniz-Zentrum fuer Informatik.

\bibitem[\protect\citename{Chakravarthi \bgroup et al.\egroup
  }2019b]{chakravarthi-etal-2019-wordnet}
Chakravarthi, B.~R., Arcan, M., and McCrae, J.~P.
\newblock (2019b).
\newblock {W}ord{N}et gloss translation for under-resourced languages using
  multilingual neural machine translation.
\newblock In {\em Proceedings of the Second Workshop on Multilingualism at the
  Intersection of Knowledge Bases and Machine Translation}, pages 1--7, Dublin,
  Ireland, August. European Association for Machine Translation.

\bibitem[\protect\citename{Chakravarthi \bgroup et al.\egroup
  }2019c]{chakravarthi-etal-2019-multilingual}
Chakravarthi, B.~R., Priyadharshini, R., Stearns, B., Jayapal, A., S, S.,
  Arcan, M., Zarrouk, M., and McCrae, J.~P.
\newblock (2019c).
\newblock Multilingual multimodal machine translation for {D}ravidian languages
  utilizing phonetic transcription.
\newblock In {\em Proceedings of the 2nd Workshop on Technologies for MT of Low
  Resource Languages}, pages 56--63, Dublin, Ireland, 20 August. European
  Association for Machine Translation.

\bibitem[\protect\citename{Chakravarthi \bgroup et al.\egroup
  }2020]{chakravarthi-etal-2020-senti-tamil}
Chakravarthi, B.~R., Muralidaran, V., Priyadharshini, R., and McCrae, J.~P.
\newblock (2020).
\newblock Corpus creation for sentiment analysis in code-mixed {Tamil-English}
  text.
\newblock In {\em Proceedings of the 1st Joint Workshop of SLTU (Spoken
  Language Technologies for Under-resourced languages) and CCURL (Collaboration
  and Computing for Under-Resourced Languages) (SLTU-CCURL 2020)}, Marseille,
  France, May. European Language Resources Association (ELRA).

\bibitem[\protect\citename{Cieliebak \bgroup et al.\egroup
  }2017]{cieliebak-etal-2017-twitter}
Cieliebak, M., Deriu, J.~M., Egger, D., and Uzdilli, F.
\newblock (2017).
\newblock A {T}witter corpus and benchmark resources for {G}erman sentiment
  analysis.
\newblock In {\em Proceedings of the Fifth International Workshop on Natural
  Language Processing for Social Media}, pages 45--51, Valencia, Spain, April.
  Association for Computational Linguistics.

\bibitem[\protect\citename{Das and
  Gamb{\"a}ck}2014]{das-gamback-2014-identifying}
Das, A. and Gamb{\"a}ck, B.
\newblock (2014).
\newblock Identifying languages at the word level in code-mixed {I}ndian social
  media text.
\newblock In {\em Proceedings of the 11th International Conference on Natural
  Language Processing}, pages 378--387, Goa, India, December. NLP Association
  of India.

\bibitem[\protect\citename{Devlin \bgroup et al.\egroup }2018]{devlin2018bert}
Devlin, J., Chang, M.-W., Lee, K., and Toutanova, K.
\newblock (2018).
\newblock {BERT}: Pre-training of deep bidirectional transformers for language
  understanding.
\newblock {\em arXiv preprint arXiv:1810.04805}.

\bibitem[\protect\citename{Diab \bgroup et al.\egroup
  }2014]{ws-2014-approaches-code}
Mona Diab, et~al., editors.
\newblock (2014).
\newblock {\em Proceedings of the First Workshop on Computational Approaches to
  Code Switching}, Doha, Qatar, October. Association for Computational
  Linguistics.

\bibitem[\protect\citename{Esuli and Sebastiani}2006]{Esuli2006sentiwordnet}
Esuli, A. and Sebastiani, F.
\newblock (2006).
\newblock Sentiwordnet: A publicly available lexical resource for opinion
  mining.
\newblock In {\em In Proceedings of the 5th Conference on Language Resources
  and Evaluation (LREC’06}, pages 417--422.

\bibitem[\protect\citename{Fellbaum}1998]{wordnet}
Fellbaum, C.
\newblock (1998).
\newblock {\em WordNet: An Electronic Lexical Database}.
\newblock Bradford Books.

\bibitem[\protect\citename{Habimana \bgroup et al.\egroup }2019]{Habimana2019}
Habimana, O., Li, Y., Li, R., Gu, X., and Yu, G.
\newblock (2019).
\newblock Sentiment analysis using deep learning approaches: an overview.
\newblock {\em Science China Information Sciences}, 63(1):111102, Dec.

\bibitem[\protect\citename{Hu and Liu}2004]{10.1145/1014052.1014073}
Hu, M. and Liu, B.
\newblock (2004).
\newblock Mining and summarizing customer reviews.
\newblock In {\em Proceedings of the Tenth ACM SIGKDD International Conference
  on Knowledge Discovery and Data Mining}, KDD ’04, page 168–177, New York,
  NY, USA. Association for Computing Machinery.

\bibitem[\protect\citename{Jiang \bgroup et al.\egroup
  }2019]{jiang-etal-2019-challenge}
Jiang, Q., Chen, L., Xu, R., Ao, X., and Yang, M.
\newblock (2019).
\newblock A challenge dataset and effective models for aspect-based sentiment
  analysis.
\newblock In {\em Proceedings of the 2019 Conference on Empirical Methods in
  Natural Language Processing and the 9th International Joint Conference on
  Natural Language Processing (EMNLP-IJCNLP)}, pages 6279--6284, Hong Kong,
  China, November. Association for Computational Linguistics.

\bibitem[\protect\citename{Jose \bgroup et al.\egroup
  }2020]{chakravarthi-code-mix-survey}
Jose, N., Chakravarthi, B.~R., Suryawanshi, S., Sherly, E., and McCrae, J.~P.
\newblock (2020).
\newblock A survey of current datasets for code-switching research.
\newblock In {\em 2020 6th International Conference on Advanced Computing \&
  Communication Systems (ICACCS)}.

\bibitem[\protect\citename{Joshi \bgroup et al.\egroup
  }2016]{joshi-etal-2016-towards}
Joshi, A., Prabhu, A., Shrivastava, M., and Varma, V.
\newblock (2016).
\newblock Towards sub-word level compositions for sentiment analysis of
  {H}indi-{E}nglish code mixed text.
\newblock In {\em Proceedings of {COLING} 2016, the 26th International
  Conference on Computational Linguistics: Technical Papers}, pages 2482--2491,
  Osaka, Japan, December. The COLING 2016 Organizing Committee.

\bibitem[\protect\citename{Kalchbrenner \bgroup et al.\egroup
  }2014]{kalchbrenner2014convolutional}
Kalchbrenner, N., Grefenstette, E., and Blunsom, P.
\newblock (2014).
\newblock A convolutional neural network for modelling sentences.
\newblock In {\em Proceedings of the 52nd Annual Meeting of the Association for
  Computational Linguistics (Volume 1: Long Papers)}, pages 655--665,
  Baltimore, Maryland, June. Association for Computational Linguistics.

\bibitem[\protect\citename{Kiela \bgroup et al.\egroup
  }2018]{kiela-etal-2018-dynamic}
Kiela, D., Wang, C., and Cho, K.
\newblock (2018).
\newblock Dynamic meta-embeddings for improved sentence representations.
\newblock In {\em Proceedings of the 2018 Conference on Empirical Methods in
  Natural Language Processing}, pages 1466--1477, Brussels, Belgium,
  October-November. Association for Computational Linguistics.

\bibitem[\protect\citename{Krishnamurti}2003]{krishnamurti2003dravidian}
Krishnamurti, B.
\newblock (2003).
\newblock {\em The Dravidian Languages}.
\newblock Cambridge University Press.

\bibitem[\protect\citename{Lalitha~Devi}2019]{lalitha-devi-2019-resolving}
Lalitha~Devi, S.
\newblock (2019).
\newblock Resolving pronouns for a resource-poor language, {M}alayalam using
  resource-rich language, {T}amil.
\newblock In {\em Proceedings of the International Conference on Recent
  Advances in Natural Language Processing (RANLP 2019)}, pages 611--618, Varna,
  Bulgaria, September. INCOMA Ltd.

\bibitem[\protect\citename{Lee and Wang}2015]{lee-wang-2015-emotion}
Lee, S. and Wang, Z.
\newblock (2015).
\newblock Emotion in code-switching texts: Corpus construction and analysis.
\newblock In {\em Proceedings of the Eighth {SIGHAN} Workshop on {C}hinese
  Language Processing}, pages 91--99, Beijing, China, July. Association for
  Computational Linguistics.

\bibitem[\protect\citename{M{\ae}hlum \bgroup et al.\egroup
  }2019]{maehlum-etal-2019-annotating}
M{\ae}hlum, P., Barnes, J., {\O}vrelid, L., and Velldal, E.
\newblock (2019).
\newblock Annotating evaluative sentences for sentiment analysis: a dataset for
  {N}orwegian.
\newblock In {\em Proceedings of the 22nd Nordic Conference on Computational
  Linguistics}, pages 121--130, Turku, Finland, September{--}October.
  Link{\"o}ping University Electronic Press.

\bibitem[\protect\citename{Mikolov \bgroup et al.\egroup
  }2013a]{mikolov2013efficient}
Mikolov, T., Chen, K., Corrado, G., and Dean, J.
\newblock (2013a).
\newblock Efficient estimation of word representations in vector space.
\newblock {\em arXiv preprint arXiv:1301.3781}.

\bibitem[\protect\citename{Mikolov \bgroup et al.\egroup }2013b]{word2vec}
Mikolov, T., Sutskever, I., Chen, K., Corrado, G.~S., and Dean, J.
\newblock (2013b).
\newblock Distributed representations of words and phrases and their
  compositionality.
\newblock In C.~J.~C. Burges, et~al., editors, {\em Advances in Neural
  Information Processing Systems 26}, pages 3111--3119. Curran Associates, Inc.

\bibitem[\protect\citename{Mohammad}2016]{mohammad-2016-practical}
Mohammad, S.
\newblock (2016).
\newblock A practical guide to sentiment annotation: Challenges and solutions.
\newblock In {\em Proceedings of the 7th Workshop on Computational Approaches
  to Subjectivity, Sentiment and Social Media Analysis}, pages 174--179, San
  Diego, California, June. Association for Computational Linguistics.

\bibitem[\protect\citename{Mouthami \bgroup et al.\egroup
  }2013]{mouthami2013sentiment}
Mouthami, K., Devi, K.~N., and Bhaskaran, V.~M.
\newblock (2013).
\newblock Sentiment analysis and classification based on textual reviews.
\newblock In {\em 2013 international conference on Information communication
  and embedded systems (ICICES)}, pages 271--276. IEEE.

\bibitem[\protect\citename{Nair \bgroup et al.\egroup }2014]{nair2014sentima}
Nair, D.~S., Jayan, J.~P., Sherly, E., et~al.
\newblock (2014).
\newblock Sentima-sentiment extraction for {M}alayalam.
\newblock In {\em 2014 International Conference on Advances in Computing,
  Communications and Informatics (ICACCI)}, pages 1719--1723. IEEE.

\bibitem[\protect\citename{Patra \bgroup et al.\egroup
  }2018]{patra2018sentiment}
Patra, B.~G., Das, D., and Das, A.
\newblock (2018).
\newblock Sentiment analysis of code-mixed indian languages: An overview of
  sail\_code-mixed shared task@ icon-2017.
\newblock {\em arXiv preprint arXiv:1803.06745}.

\bibitem[\protect\citename{Pennington \bgroup et al.\egroup
  }2014]{pennington2014glove}
Pennington, J., Socher, R., and Manning, C.~D.
\newblock (2014).
\newblock Glove: Global vectors for word representation.
\newblock In {\em EMNLP}, volume~14, pages 1532--1543.

\bibitem[\protect\citename{{Poria} \bgroup et al.\egroup }2012]{sentinet}
{Poria}, S., {Gelbukh}, A., {Cambria}, E., {Yang}, P., {Hussain}, A., and
  {Durrani}, T.
\newblock (2012).
\newblock {Merging SenticNet and WordNet-Affect emotion lists for sentiment
  analysis}.
\newblock In {\em 2012 IEEE 11th International Conference on Signal
  Processing}, volume~2, pages 1251--1255, Oct.

\bibitem[\protect\citename{Priyadharshini \bgroup et al.\egroup
  }2020]{chakravarthi-code-mix-ruba-ne}
Priyadharshini, R., Chakravarthi, B.~R., Vegupatti, M., and McCrae, J.~P.
\newblock (2020).
\newblock Named entity recognition for code-mixed {I}ndian corpus using meta
  embedding.
\newblock In {\em 2020 6th International Conference on Advanced Computing \&
  Communication Systems (ICACCS)}.

\bibitem[\protect\citename{Rani \bgroup et al.\egroup
  }2020]{priya-etal-2020-senti-comparative}
Rani, P., Suryawanshi, S., Goswami, K., Chakravarthi, B.~R., Fransen, T., and
  McCrae, J.~P.
\newblock (2020).
\newblock A comparative study of different state-of-the-art hate speech
  detection methods for {Hindi-English} code-mixed data.
\newblock In {\em Proceedings of the Second Workshop on Trolling, Aggression
  and Cyberbullying}, Marseille, France, May. European Language Resources
  Association (ELRA).

\bibitem[\protect\citename{{Ranjan} \bgroup et al.\egroup
  }2016]{chakravarthi2016}
{Ranjan}, P., {Raja}, B., {Priyadharshini}, R., and {Balabantaray}, R.~C.
\newblock (2016).
\newblock A comparative study on code-mixed data of {I}ndian social media vs
  formal text.
\newblock In {\em 2016 2nd International Conference on Contemporary Computing
  and Informatics (IC3I)}, pages 608--611, Dec.

\bibitem[\protect\citename{Rogers \bgroup et al.\egroup
  }2018]{rogers-etal-2018-rusentiment}
Rogers, A., Romanov, A., Rumshisky, A., Volkova, S., Gronas, M., and Gribov, A.
\newblock (2018).
\newblock {R}u{S}entiment: An enriched sentiment analysis dataset for social
  media in {R}ussian.
\newblock In {\em Proceedings of the 27th International Conference on
  Computational Linguistics}, pages 755--763, Santa Fe, New Mexico, USA,
  August. Association for Computational Linguistics.

\bibitem[\protect\citename{Rosowsky}2010]{rosowsky2010writing}
Rosowsky, A.
\newblock (2010).
\newblock {‘Writing it in English’: script choices among young multilingual
  Muslims in the UK}.
\newblock {\em Journal of Multilingual and Multicultural Development},
  31(2):163--179.

\bibitem[\protect\citename{Saint-Jacques}1987]{saint1987roman}
Saint-Jacques, B.
\newblock (1987).
\newblock {The Roman alphabet in the Japanese writing system}.
\newblock {\em Visible Language}, 21(1):88.

\bibitem[\protect\citename{Sarkar and Chakraborty}2015]{sarkar2015sentiment}
Sarkar, K. and Chakraborty, S.
\newblock (2015).
\newblock A sentiment analysis system for indian language tweets.
\newblock In {\em International Conference on Mining Intelligence and Knowledge
  Exploration}, pages 694--702. Springer.

\bibitem[\protect\citename{Scotton}1982]{scotton1982possibility}
Scotton, C.~M.
\newblock (1982).
\newblock The possibility of code-switching: motivation for maintaining
  multilingualism.
\newblock {\em Anthropological linguistics}, pages 432--444.

\bibitem[\protect\citename{Se \bgroup et al.\egroup }2015]{se2015amrita}
Se, S., Vinayakumar, R., Kumar, M.~A., and Soman, K.
\newblock (2015).
\newblock {AMRITA-CEN@ SAIL2015: sentiment analysis in Indian languages}.
\newblock In {\em International Conference on Mining Intelligence and Knowledge
  Exploration}, pages 703--710. Springer.

\bibitem[\protect\citename{Se \bgroup et al.\egroup }2016]{se2016predicting}
Se, S., Vinayakumar, R., Kumar, M.~A., and Soman, K.
\newblock (2016).
\newblock Predicting the sentimental reviews in {T}amil movie using machine
  learning algorithms.
\newblock {\em Indian Journal of Science and Technology}, 9(45).

\bibitem[\protect\citename{Severyn \bgroup et al.\egroup
  }2014]{severyn-etal-2014-opinion}
Severyn, A., Moschitti, A., Uryupina, O., Plank, B., and Filippova, K.
\newblock (2014).
\newblock Opinion mining on {Y}ou{T}ube.
\newblock In {\em Proceedings of the 52nd Annual Meeting of the Association for
  Computational Linguistics (Volume 1: Long Papers)}, pages 1252--1261,
  Baltimore, Maryland, June. Association for Computational Linguistics.

\bibitem[\protect\citename{Solorio \bgroup et al.\egroup
  }2014]{solorio-etal-2014-overview}
Solorio, T., Blair, E., Maharjan, S., Bethard, S., Diab, M., Ghoneim, M.,
  Hawwari, A., AlGhamdi, F., Hirschberg, J., Chang, A., and Fung, P.
\newblock (2014).
\newblock Overview for the first shared task on language identification in
  code-switched data.
\newblock In {\em Proceedings of the First Workshop on Computational Approaches
  to Code Switching}, pages 62--72, Doha, Qatar, October. Association for
  Computational Linguistics.

\bibitem[\protect\citename{Sreelekha and Bhattacharyya}2018]{S18.125}
Sreelekha, S. and Bhattacharyya, P.
\newblock (2018).
\newblock {Morphology Injection for English-Malayalam Statistical Machine
  Translation}.
\newblock In Nicoletta Calzolari~(Conference chair), et~al., editors, {\em
  Proceedings of the Eleventh International Conference on Language Resources
  and Evaluation (LREC 2018)}, Miyazaki, Japan, May 7-12, 2018. European
  Language Resources Association (ELRA).

\bibitem[\protect\citename{Suryawanshi \bgroup et al.\egroup
  }2020a]{suryawanshi-etal-2020-meme}
Suryawanshi, S., Chakravarthi, B.~R., Arcan, M., and Buitelaar, P.
\newblock (2020a).
\newblock Multimodal meme dataset ({MultiOFF}) for identifying offensive
  content in image and text.
\newblock In {\em Proceedings of the Second Workshop on Trolling, Aggression
  and Cyberbullying}, Marseille, France, May. European Language Resources
  Association (ELRA).

\bibitem[\protect\citename{Suryawanshi \bgroup et al.\egroup
  }2020b]{suryawanshi-etal-2020-tamil-meme}
Suryawanshi, S., Chakravarthi, B.~R., Verma, P., Arcan, M., McCrae, J.~P., and
  Buitelaar, P.
\newblock (2020b).
\newblock A dataset for troll classification of {Tamil} memes.
\newblock In {\em Proceedings of the 5th Workshop on Indian Language Data
  Resource and Evaluation (WILDRE-5)}, Marseille, France, May. European
  Language Resources Association (ELRA).

\bibitem[\protect\citename{Tay}1989]{tay1989code}
Tay, M.~W.
\newblock (1989).
\newblock Code switching and code mixing as a communicative strategy in
  multilingual discourse.
\newblock {\em World Englishes}, 8(3):407--417.

\bibitem[\protect\citename{Thottingal}2019]{thottingal-2019-finite}
Thottingal, S.
\newblock (2019).
\newblock Finite state transducer based morphology analysis for {M}alayalam
  language.
\newblock In {\em Proceedings of the 2nd Workshop on Technologies for MT of Low
  Resource Languages}, pages 1--5, Dublin, Ireland, August. European
  Association for Machine Translation.

\bibitem[\protect\citename{Valitutti}2004]{Valitutti04wordnet-affect:an}
Valitutti, R.
\newblock (2004).
\newblock {WordNet-Affect}: an affective extension of wordnet.
\newblock In {\em In Proceedings of the 4th International Conference on
  Language Resources and Evaluation}, pages 1083--1086.

\bibitem[\protect\citename{Wiebe \bgroup et al.\egroup }2005]{Wiebe2005}
Wiebe, J., Wilson, T., and Cardie, C.
\newblock (2005).
\newblock Annotating expressions of opinions and emotions in language.
\newblock {\em Language Resources and Evaluation}, 39(2):165--210, May.

\bibitem[\protect\citename{Zhou \bgroup et al.\egroup }2016]{zhou2016text}
Zhou, P., Qi, Z., Zheng, S., Xu, J., Bao, H., and Xu, B.
\newblock (2016).
\newblock Text classification improved by integrating bidirectional lstm with
  two-dimensional max pooling.
\newblock {\em arXiv preprint arXiv:1611.06639}.

\end{thebibliography}

\end{document}